%% file: reduced_version_ICRA21.tex
\documentclass[letterpaper, 10 pt, conference]{ieeeconf}  
\IEEEoverridecommandlockouts                 
\usepackage{graphicx}
\usepackage{amsmath,amssymb}
\usepackage{ntheorem}
\usepackage{mathrsfs}  \usepackage{color,soul}
\usepackage{enumerate}

\usepackage[ruled,linesnumbered]{algorithm2e}
\usepackage{algpseudocode}
\usepackage{cite} 
\usepackage[dvipsnames, table]{xcolor} 

\newcommand{\conn}{\lambda_2}

\newcommand{\cbf}{\conn(x) - \varepsilon}
\DeclareMathOperator*{\argmin}{argmin}

\usepackage{nicefrac}
\usepackage{paralist}
\usepackage[normalem]{ulem}
\usepackage{multirow}  
\usepackage{booktabs}  
\usepackage{array}
\usepackage{diagbox}  
\usepackage{siunitx}

\usepackage{tikz}
\usepackage{pgfplots}	
\usepgfplotslibrary{groupplots}
\usepackage{siunitx}
\usepackage{pgfplotstable}
\usepackage[caption=false]{subfig} 

\usepackage{todonotes} 

\DeclareCaptionType{Table}
 
\graphicspath{{images/}}
 
\definecolor{color1}{rgb}{0,0.4470,0.7410}
\definecolor{color2}{rgb}{0.8500,0.3250,0.0980}
\definecolor{color3}{rgb}{0.9290,0.6940,0.1250}
\definecolor{color4}{rgb}{0.4940,0.1840,0.5560}
\definecolor{color5}{rgb}{0.4660,0.6740,0.1880}
\definecolor{color6}{RGB}{19,122,16}

\pgfplotscreateplotcyclelist{MyCyclelist}{%
	{color1, mark = none, line width=0.6pt},
	{color3, mark = none, line width=0.6pt},
	{color4, mark = none, line width=0.6pt},
	{color5, mark = none, line width=0.6pt},
	{color4, mark = none, line width=0.6pt}
}

\definecolor{my_orange}{RGB}{255,121,3}
\definecolor{my_light_green}{RGB}{32,255,42}
\definecolor{my_yellow}{RGB}{255,255,32}
\definecolor{my_light_blue}{RGB}{50,255,255}
\definecolor{my_blue}{RGB}{50,120,255}

\newcolumntype{C}[1]{>{\centering}m{#1}} 

\begin{document}
\title{\LARGE \bf
	Decentralized Connectivity Maintenance with Time Delays using Control Barrier Functions}


\author{Beatrice Capelli$^1$, Hassan Fouad$^2$, Giovanni Beltrame$^2$, Lorenzo Sabattini$^1$%
	%
	%
	%
	\thanks{$^1$ Beatrice Capelli and Lorenzo Sabattini are with the Department of Sciences and Methods for Engineering (DISMI), University of Modena and Reggio Emilia, Italy {\tt\small{\{beatrice.capelli, lorenzo.sabattini\}@unimore.it}}} 
	\thanks{$^2$ Hassan Fouad and Giovanni Beltrame are with the Department of Computer and Software Engineering, \'Ecole Polytechnique de Montr\'eal, Qu\'ebec, Canada {\tt\small{\{hassan.fouad, giovanni.beltrame\}@polymtl.ca}}}%
}

%

\maketitle

\begin{abstract}                          
  Connectivity maintenance is crucial for the real world deployment of multi-robot
  systems, as it ultimately allows the robots to communicate, coordinate and perform tasks in
  a collaborative way. 
  A connectivity maintenance controller must keep the multi-robot
  system connected independently from the system's mission and in the
  presence of undesired real world effects such as communication
  delays, model errors, and computational time delays, among others.
  In this paper we present the implementation, on a real robotic
  setup, of a connectivity maintenance control strategy based
  on 
  Control Barrier Functions. 
  During experimentation, we found that the presence of communication
  delays has a significant impact on the performance of the controlled
  system, with respect to the ideal case.
%
  We propose a heuristic to counteract the effects of communication
  delays, and we verify its efficacy both in simulation and with
  physical robot experiments.

\end{abstract}


\section{Introduction}
The interest for multi-robot systems is constantly increasing, in a
%
wide range of fields, from industrial~\cite{ram2017}, to
agricultural~\cite{ball2015}, marine~\cite{kemna2018}, and
aerial~\cite{saska2017} applications.
To be able to collaborate, robots must be able to communicate, and a
controller that maintains connectivity is greatly beneficial to the
implementation of multi-robot applications.

In literature, the connectivity maintenance problem is usually
addressed from two different points of view: local and global. The
local approach~\cite{ji2007distributed, dimarogonas2008decentralized,
  ajorlou2010class} preserves the local connections among the robots,
resulting in a connected system, in a sort of bottom-up
approach. Conversely, global connectivity~\cite{yang2010decentralized,
  sabattini2013decentralized, li2013bounded} considers the overall
robot network in a top-down approach. A detailed comparison between
the two approaches can be found in~\cite{khateri2019comparison}.

A multi-robot system must also be able to perform a given task in an
efficient way, on top of keeping connectivity. For this purpose,
\cite{capelli2020cbf} presents a Control Barrier Function that meets
both requirements. Control Barrier Functions
(CBFs)~\cite{ames2019control} are a control technique that allows a
system to simultaneously achieve its objectives while maintaining some
constraints. The core of the approach is that of a minimally intrusive
controller with respect to the desired one. The desired control can be
generated according to the robots' mission, e.g.,
coverage~\cite{cortes2010coverage}, formation
control~\cite{sabattini2011arbitrarily},
flocking~\cite{vasarhelyi2018optimized} or
patrolling~\cite{pasqualetti2012cooperative}. For multi-robot systems,
examples of constraints include energy
persistence~\cite{notomista2018persistification}, collision
avoidance~\cite{wang2017safety}, local~\cite{wang2016multi} and global
connectivity~\cite{capelli2020cbf} among others.

In~\cite{wang2016safety, wang2017safety} a decentralized approach was
introduced, but the implementation was carried out in a centralized way,
namely all the calculations were done on a central unit.

When addressing a real deployment, one must take into account time
delays. It is well known that delays usually lead to instability in
controlled systems, if they are not properly
considered~\cite{secchi2007control}. Some recent
works~\cite{jankovic2018control, orosz2019safety} addressed the delay
problem in the CBF approach. In particular,
in~\cite{jankovic2018control} time delays were approximated and the
proposed solution consisted in using the predicted value of the state
for the CBF, instead of the current one. In~\cite{orosz2019safety} the
time delays were not approximated and the existence of safety
functionals was investigated, but the CBF was not explicitly derived
due to the complexity of the problem.


The contribution of this paper is the definition of a heuristic method
to implement a CBF-based control strategy on a decentralized
multi-robot setup, affected by the presence of communication time
delays. Building upon the results in~\cite{capelli2020cbf}, we define
a control strategy that guarantees connectivity with good performance
even in the presence of communication delays.


The paper is organized as follows. Preliminary notions related
to graph theory and CBFs are provided in
Section~\ref{sec:background}. Section~\ref{sec:system_definition}
reports the system definition and the problem statement. In
Section~\ref{sec:heuristic} we introduce the proposed heuristic for
the implementation of CBFs in systems that suffer from delays. The
results of the experiments and the conclusions are reported in
Section~\ref{sec:experimental_validation} and
Section~\ref{sec:conclusion}, respectively.




\section{Background}
\label{sec:background}
In this section we introduce the two main theoretical instruments that we use
in this work: connectivity, derived from graph theory, and
CBFs. 

\subsection{Notation}
\label{sec:notation}
$\mathbb{R}$, $\mathbb{R}^+_0$, $\mathbb{R}^+$ are the set of real, real
non-negative, and real positive numbers, respectively. The set of locally
Lipschitz functions is $\mathcal{L}$. A continuous
function $\Omega(\cdot): \mathbb{R}^+_0 \rightarrow \mathbb{R}^+_0$ is a class
$\mathcal{K}$ function if it is strictly increasing and $\Omega(0) = 0$. It is
an extended class $\mathcal{K}$ function if it is a class $\mathcal{K}$ function, and
it is defined on the entire real line, e.g.,
$\Omega(\cdot): \mathbb{R} \rightarrow \mathbb{R}$~\cite{khalil2002nonlinear}.

\subsection{Connectivity}
\label{sec:conn}
Graph theory is usually used to represent the communication topology
of a multi-robot system. In particular, the set of robots are
represented as a set of vertices $V$ and a set of edges $E$ based on
the type of communication model (e.g., R-disk, line of sight,
etc.). The set $E$ consists of the edges $e_{i,j}$ between two robots
$i$ and $j$ that are able to communicate. In this paper we consider an
R-disk communication model, and hence two robots can communicate if
and only if they are within the communication distance $R$. In
addition, we consider an undirected graph, so if the edge $e_{i,j}$
exists, then also the edge $e_{j,i}$ exists. The overall communication
topology is represented by the communication graph
$\mathcal{G} = \{V,E\}$.

For control purposes, it is important to quantify the connectivity
status of the system. The most common
measure~\cite{yang2010decentralized} used for this purpose is the
algebraic connectivity or Fiedler value~\cite{fiedler1973algebraic},
$\conn$. It represents the sparsity of the graph, and if
$\conn > 0$, then the graph is connected. 

Algebraic connectivity is the second smallest eigenvalue of the
Laplacian matrix~\cite{mohar1991laplacian}. The Laplacian matrix is
defined as $L = D - A$, where $D$ is the degree matrix and $A$ is the
adjacency matrix. If we consider a group of $N$ robots, and defining
the neighbors of the $i$-th robot
\mbox{$\mathcal{N}_i = \{ j \in V | e_{i,j} \in E \}$}, we can define the
adjacency matrix $A \in \mathbb{R}^{N \times N}$ as:
\begin{equation}
A = \begin{cases}
a_{i,j} > 0 & \text{if } j \in \mathcal{N}_i \\
0 & \text{otherwise}
\end{cases}
\end{equation}
where $a_{i,j}$ is the edge weight of $e_{i,j}$. The degree matrix $D \in \mathbb{R}^{N \times N}$ is a diagonal matrix $D = diag\{\psi_{i,i}\}$, where $\psi_{i,i} = \sum_{j=1}^{N}a_{i,j}$.

\subsection{Control Barrier Functions} 
We want to consider connectivity maintenance as a constraint over the
general goal of a multi-robot system.
CBFs 
are a suitable tool for this kind of problem, generating a control
input that is minimally intrusive with respect to the ideal one.
Consider the affine control system:
\begin{equation}
\label{eq:affine}
\dot{\chi} = f(\chi) + g(\chi)\mu
\end{equation}
where $\chi \in \mathbb{R}^p$ represents the state of the system,
\mbox{$\mu \in U \subseteq \mathbb{R}^q$} is the control input, with
$U$ defined as the set of admissible inputs for the system. Moreover,
we assume $f(\chi), g(\chi) \in \mathcal{L}$.

Now consider a desired constraint that can be expressed as a
superlevel set of a continuously differentiable function
$h(\chi) : \mathbb{R}^p \rightarrow \mathbb{R}$. The superlevel set
$\mathcal{C} \subset \mathbb{R}^p$ is called the safety set for the
system and it is defined as: \mbox{$\mathcal{C} = \left\{\chi \in \mathbb{R}^p | h(\chi) \geq 0\right\}$}.
The objective of the CBF is to keep the system inside the safety set,
i.e., to render the set $\mathcal{C}$ forward invariant. A set
$\mathcal{C}$ is forward invariant if, for every
$\chi_0 \in \mathcal{C}$, then $\chi(t) \in \mathcal{C}$ for
$\chi(0) = \chi_0$ and $\forall t > 0$. The system~\eqref{eq:affine}
is safe with respect to the set $\mathcal{C}$ if the set $\mathcal{C}$
is forward invariant.

The function
$h(\chi): \mathcal{D} \subset \mathbb{R}^p \rightarrow \mathbb{R}$ is
a CBF if there exists an extended class $\mathcal{K}$ function
$\alpha\left(h(\chi)\right)$ such that~\cite{ames2019control}:
\begin{equation}
\label{eq:sup}
\sup\limits_{\mu \in U} \left[ L_f h(\chi) + L_g h(\chi)\mu + \alpha\left(h(\chi)\right) \right] \geq 0, \text{ } \forall \chi \in \mathcal{D}
\end{equation} 
where $L_f$ and $L_g$ represent the Lie derivatives of $h(x)$: \mbox{$L_f h(x) = \frac{\partial h(x)}{\partial x} f(x)$,} $L_g h(x) = \frac{\partial h(x)}{\partial x} g(x)$.
In order to render the system~\eqref{eq:affine} safe with respect to the desired set $\mathcal{C}$, the control input $\mu(\chi): \mathcal{D} \rightarrow U \text{, } \mu(\chi) \in \mathcal{L}$ must belong to the set $K_{cbf}(\chi)$, which is defined as:
\mbox{$K_{cbf}(\chi) = \{ \mu \in U | L_f h(\chi) + L_g h(\chi)\mu + \alpha\left(h(\chi)\right) \geq 0 \}$}.

\section{System definition and problem statement}
\label{sec:system_definition}

\subsection{System dynamics and control strategy}

Consider a system of $N$ robots that are able to move in a
$n$-dimensional space. In the case of ground robots $n=2$ (as in the
experiments reported hereafter), while in the case of aerial robots
$n=3$. We define the state of the system
$x = \left[ x_1^T, \dotsc, x_N^T \right]^T \in \mathbb{R}^{nN}$, where
$x_i \in \mathbb{R}^n$ represents the position of the $i$-th robot.

In addition, we assume a single integrator dynamics:
\begin{equation}
    \label{eq:sinint}
    \dot{x} = u
\end{equation}
where $u \in U \subseteq \mathbb{R}^{nN}$ represents the control input of the
system.  It is worth
remarking that, by using a sufficiently good
Cartesian trajectory tracking controller, it is possible to represent the kinematic behavior of several
types of mobile robots, like wheeled mobile robots~\cite{soukieh2009obstacle}, and
UAVs~\cite{lee2013semiautonomous}, with~\eqref{eq:sinint}. We can now instantiate the general affine
control system, reported in~\eqref{eq:affine}, with
$f(x) = \mathbb{O} \in \mathbb{R}^{nN \times nN}$,
\mbox{$g(x) = \mathbb{I} \in \mathbb{R}^{nN\times nN}$}. $\mathbb{O}$ and $\mathbb{I}$
represent, respectively, the null and the identity matrix of opportune dimension.

Along the lines of~\cite{tro2017, capelli2020cbf}, we calculate the
edge weights of the graph $\mathcal{G}$ as a function of the Euclidean
distance between the $i\text{-th}$ and the $j$-th robot,
$d_{i,j} = \| x_i - x_j \|$. Considering the communication distance
$R \in \mathbb{R}^+$, and introducing a constant
$\sigma\in \mathbb{R}^+$ for normalization purpose\footnote{One
  possible choice for $\sigma$ is $\frac{R^4}{log(2)}$, in order to
  obtain a value $a_{i,j} \leq 1$.}, we define:
\begin{equation}
    \label{eq:weight}
    a_{i,j} = \begin{cases}
    e^{\left(R^2 - d_{i,j}^2\right)^2/\sigma} -1 & \text{if\ } d_{i,j} \leq R \\
    0 & \text{otherwise}
    \end{cases}
\end{equation}

In~\cite{capelli2020cbf}, the following CBF was introduced for connectivity
maintenance:
\begin{equation}
    \label{eq:cbf}
    h(x) = \cbf
\end{equation}
where $\conn(x)$ is the algebraic connectivity of the system and
$\varepsilon \in \mathbb{R}^+$ is an adjustable threshold to allow different
levels of action of the constraint, namely the higher $\varepsilon$, the
higher the desired global connectivity value. The proposed
CBF~\eqref{eq:cbf} was verified in simulation to be effective in
multiple scenarios, that is considering different desired inputs $u_{des}$.

The general solution for this problem is obtained solving the
following Quadratic Program (QP):
\begin{equation}
    \label{eq:qp}
    \begin{aligned}
        u(x) = \argmin\limits_{u \in \mathbb{R}^{nN}} \quad & \frac{1}{2}\| u - u_{des}(x) \|^2\\
        \text{s.t. } \quad & \frac{\partial \lambda_2}{\partial x}u \geq - \left(\lambda_2 -\varepsilon \right)\\
        & u \in U
    \end{aligned}
\end{equation}
where we set the extended class $\mathcal{K}$ function $\alpha(\cdot)$
equal to the identity, i.e., $\alpha(h(x)) = h(x)$.

The QP problem~\eqref{eq:qp} can be solved in a decentralized fashion
if each robot solves the problem considering the $i$-th component of
the first constraint. To compute $\frac{\partial \conn}{\partial x_i}$
in a distributed way, we use the the results
of~\cite{yang2010decentralized}, from which:
\begin{equation}
    \label{eq:der_conn}
    \frac{\partial \conn}{\partial x_i}
    = \sum_{j \in \mathcal{N}_i}  \frac{\partial a_{i,j}}{\partial x_i} \left(v_2^i - v_2^j\right)^2  
\end{equation}
where $v_2^i$ and $v_2^j$ are the $i$-th and $j$-th component of the
eigenvector associated to $\conn$, respectively.  We enable the robots
to communicate their reciprocal distances to the whole group, in such
a way that they can build the Laplacian matrix, and from it they can
calculate the components needed in~\eqref{eq:der_conn}.

To avoid collisions, a necessary condition from~\cite{capelli2020cbf},
we used the CBF proposed in~\cite{egerstedt2018robot}:
\begin{equation}
\label{eq:cbf_collision}
h_{safety}(x_i, x_j) = d_{i,j}^2 - d_{min}^2
\end{equation}
where $d_{min} \in \mathbb{R}^+$ is the minimum safe distance between
robots. The reader can refer to~\cite{glotfelter2017nonsmooth,
  egerstedt2018robot} for a detailed description of the composition of
CBFs.

\subsection{Problem statement: decentralized implementation}
We consider the problem of implementing, in a fully distributed
setting, the CBF-based control strategy on a team of robots. In
particular, we consider the following issues:
\begin{enumerate}
\item \emph{Decentralized implementation}: each robot computes its control
  input independently, based on the data available on board;
\item \emph{Imperfect state knowledge}: each robot can measure its own state,
  and can communicate it to the rest of the robots in the group. Each robot
  has then access to the full state of the multi-robot system, but the
  information is subject to communication delays.
\end{enumerate}

\section{Heuristic for CBF implementation in the presence of delays}\label{sec:heuristic}
As reported in Section~\ref{sec:preliminary_experiments}, we performed
initial experiments to evaluate the behavior of the control strategy
proposed in~\cite{capelli2020cbf}. When using a decentralized
multi-robot setup, the presence of communication delays cause a
degradation of the system performance.

Considering a constant delay $\tau \in \mathbb{R}^+$, the general
model of the system given in~\eqref{eq:affine} can be rewritten as
\begin{equation}
\label{eq:affine_delay}
\dot{x}(t) = f\left(x(t), x(t - \tau) \right) + g(x(t), x(t - \tau))u(t)
\end{equation}
to explicitly take into account the delayed state information used by
each robot.

As reported in~\cite{orosz2019safety}, the presence of such delay
modifies the safety set of the CBF. However, the explicit computation
of this modified safety set is challenging, and not practical in
the general case. To avoid the need to explicitly compute the modified
safety set, we propose a heuristic.

Each robot aggregates the state $x$ from all other robots in the network
through multi-hop communication, a process that introduces time delay due to
communication latency. From the state, each robot can calculate the value of
$\conn$, which, in turn, is affected by the delay. More specifically, the
computation taken at time $t$ is based on the state
$x\left(t-\tau\right)$. Hence, the computed value of $\conn$ does not
represent the \emph{real} connectivity condition, which should use the
state $x(t)$.

\begin{figure}[tb]
	\centering
	\def\svgwidth{0.6\columnwidth}
	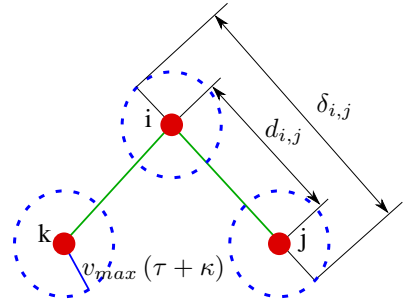
	\caption{Schematic for the proposed heuristic. In \textit{red} the robots, in \textit{blue} the possible area in which each robot can be after the time $\tau$, and in \textit{green} the current edges.}
	\label{fig:heuristic}
\end{figure}

It is worth remarking that $\conn$ is a non-decreasing function of
each edge weight. Considering the definition given
in~\eqref{eq:weight}, the value of each edge weight $a_{i,j}$
decreases, as the inter-robot distance $d_{i,j}$ increases. Hence, we
propose a heuristic based on the worst-case scenario that takes place
when, during the time $\tau$, the increase of the distance $d_{i,j}$
is maximum. Namely, when robots $i$ and $j$ move in opposite
directions at the maximum of their velocity. Fig.~\ref{fig:heuristic}
shows a schematic representation of the idea behind the proposed
heuristic.

Hence, the proposed heuristic consists of redefining the edge weights
in~\eqref{eq:weight}, replacing the measured distance $d_{i,j}$ with a
modified distance $\delta_{i,j}$, defined as
%
%
\begin{equation}
\label{eq:heuristic}
\delta_{i,j}= d_{i,j} +  2 v_{max} \left( \tau + \kappa \right)
\end{equation}
where $v_{max}$ is the maximum velocity of the robots. We also
introduce a correction factor $\kappa \in \mathbb{R}^+$
to take into account additional disturbing factors that can affect the
computation of $\conn$ from the communicated state. These disturbing
factors include all the unmodelled elements that characterize the real
system with respect to the ideal one, such as real robot kinematics
with limited inputs, model errors, etc.

\section{Experimental validation}
\label{sec:experimental_validation}
In this section we describe the experiments we carried out to
demonstrate the effectiveness of the proposed control strategy, in the
presence of time delays.

\subsection{Hardware and software implementation}
\label{sec:hardware_software_implementation}
The experimental validation of the proposed connectivity maintenance
strategy had been carried out on a group of K-Team
Khepera IV (KH4)
robots deployed inside a $2\times2 \;\si{\meter}$ arena equipped with
an OptiTrack tracking system. The Khepera robot is equipped a with
800MHz ARM Cortex-A8 processor, mounting the Yocto operating
system. The robots communicate through Wi-Fi and a software hub,
\texttt{blabbermouth}
which emulates range and bearing sensors, namely each robot has the
(simulated) ability to measure the distance and orientation between
itself and its neighbors. In addition, \texttt{blabbermouth} can
emulate limited communication range, packet drops, etc.

\begin{figure}[tb]
	\centering
	\subfloat[][\centering Experimental setup.]{{\includegraphics[width=.48\linewidth]{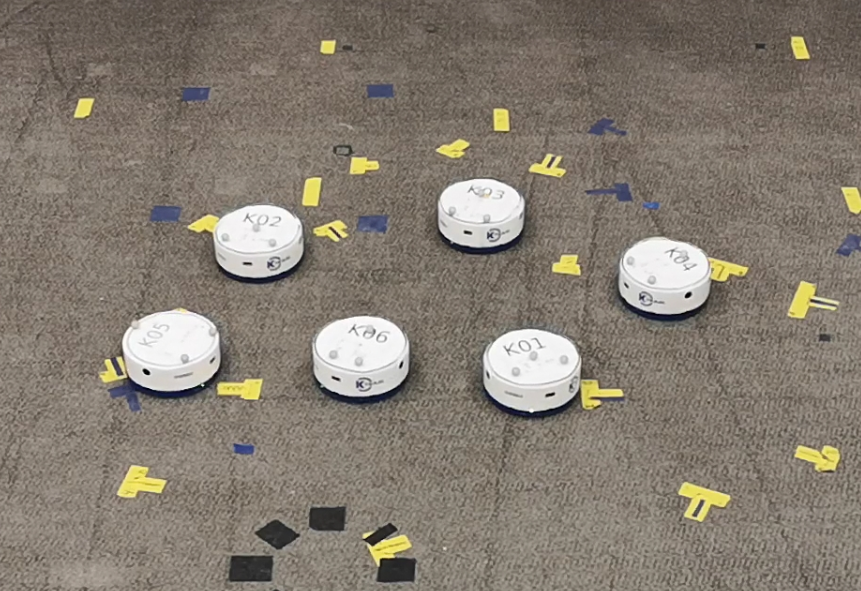} \label{fig:real_setup}}}\hfill
	\subfloat[][\centering ARGoS simulator.]{{\includegraphics[width=.48\linewidth]{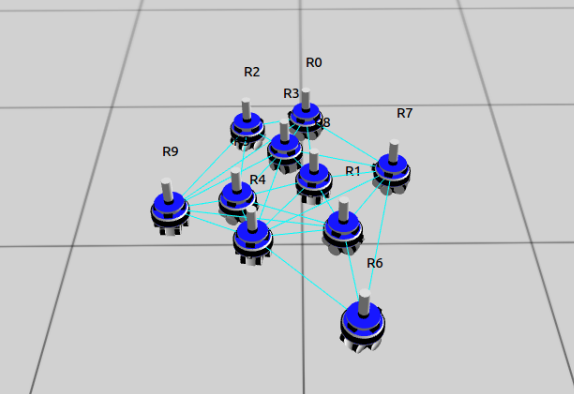} \label{fig:argos_simulation}}}%
	\caption{Experimental validation.}%
	\label{fig:experimental_setup}%
\end{figure}

The control algorithm is implemented in Buzz~\cite{pinciroli2016swarm}, which
is a programming language specific for swarm behaviors. In particular, during
the experiments, we use the virtual stigmergy~\cite{pinciroli2016tuple}, a
mechanism to share information through the multi-robot system. This mechanism
allowed us to distribute the relative positions among all the robots. From
this information, each robot could compute the value of $\conn$ and of the
corresponding eigenvector of the Laplacian matrix ($v_2$). These values were
used to define the QP problem~\eqref{eq:qp} solved on board of each robot,
with the \texttt{alglib} library.


For our simulations, we used ARGoS~\cite{pinciroli2012argos}, a
physics-based multi-robot simulator. ARGoS supports Buzz and allowed
us to use the same scripts in simulation and on the real robots.

We did not use single-integrator dynamics~\eqref{eq:sinint} directly: the
robots have differential-drive kinematics that need an additional
transformation of the velocity inputs. We used input-output state feedback
linearization~\cite{siciliano2010robotics} to transform the velocity input $u$
into the suitable velocity commands for the robots, simulated and real.

We tested the CBF for connectivity maintenance with different desired
behaviors. Due to space limitations, in the following we report only
the results of the two most relevant experiments. Some representative
runs of the experiments are shown in the attached video.  
For both the simulations and the real implementation we set: $R=1\si{\meter}$, $d_{min}=0.25\si{\m}$, and $u_{max}=0.2\si{\m / s}$.

\subsection{Initial experiments}
\label{sec:preliminary_experiments}
Our
first experiments implement the control strategy proposed
in~\cite{capelli2020cbf} on the robots. It is worth noting that the
simulation results reported in~\cite{capelli2020cbf} were carried out
in an ideal case, while the experiments in this paper consider a more
realistic situation. Table~\ref{tab:simulations_vs_experiments}
summarizes the differences.

\begin{table}[tb] 
	\centering
	\caption{Main differences between previous simulations, reported in~\cite{capelli2020cbf}, and the experiments of this paper.}
	\label{tab:simulations_vs_experiments} 
	\begin{tabular}{ ccc } 
		\toprule
		 & Previous simulations~\cite{capelli2020cbf} & Experiments \\
		\midrule
		Implementation & Centralized & Decentralized \\
		\midrule
		Positions & Absolute & Relative \\
		\midrule
		Kinematics & Omni-directional & Differential-drive \\
		\midrule
		Input & $u \in \mathbb{R}^{nN}$ & $u \in U $ \\
		\bottomrule
	\end{tabular}
\end{table}
The first desired behavior was the most challenging for the
connectivity issue: the robots were controlled to disconnect (disconnecting behavior).
Namely, the robots run away from each other, with a desired controller defined
for the $i$-th robot, where $i \in [1, \dotsc, N]$, as:
\begin{equation}
    \label{eq:disgregative}
    u_{des}^i = \left[ 
    k \cos \left( \frac{2 \pi}{N + 1} i\right) \quad
    k \sin \left( \frac{2 \pi}{N + 1} i\right)
    \right]^T
\end{equation}
where $k \in \mathbb{R}^+$ is a tuning parameter.

Fig.~\ref{fig:disconnecting_ideal_matlab} illustrates the behavior of
$\conn$ in an ideal simulation, similar to the ones presented
in~\cite{capelli2020cbf}.

This control would have led to disconnection, but, as can be
seen from Fig.~\ref{fig:relative_real}, the proposed method reacts and
prevents disconnections also in the real experiments. However, the threshold value is exceeded, namely
$\conn < \varepsilon$. It is worth noting that the oscillating behavior is caused by the non-holonomic kinematics of the robots. To reduce this issue we can add an artificial damping to the dynamics of the robots.

\begin{figure}[tb]	
	\centering
	\begin{subfloat}[][\centering Ideal conditions, as in the previous work~\cite{capelli2020cbf} ($N=10$, $\varepsilon=0.3$).]{
	\begin{tikzpicture}
		\begin{axis}[
			height = 4cm,
			width = 0.5\columnwidth,
			xlabel near ticks, 
			xlabel = {t [\si{\second}]},
			ylabel near ticks, 
			ylabel shift={-4pt},
			ylabel = {$\conn$ },
			ymin=0, ymax=2,
			ytick={0, 1, 2, 3, 4},
			yticklabels = {$0$, $1$, $2$, $3$, $4$},
			xmin=0, xmax=49,
			xtick={0,25, 49},
			xticklabels = {$0$, $25$, $50$},,
			legend image post style={only marks, mark=-},
			legend style={font=\footnotesize}
			]
			\addplot [line width=1pt, color=red] 
			table[x = x_pos, y = y_pos,col sep=space, row sep=\\]
			{
				x_pos	y_pos\\
				0  0.3\\
				500  0.3\\
			};
			\addplot [line width=1.4pt, color=color1] table [x index=1,y index=0,col sep=comma] {data/disconnecting_ideal_matlab.csv};	
			\legend{$\varepsilon = 0.3$}
		\end{axis}
	\end{tikzpicture}
	\label{fig:disconnecting_ideal_matlab}}
	\end{subfloat}
	\begin{subfloat}[][\centering Real conditions in initial experiments on real robots ($N=3$, $\varepsilon=0.3$).]{
		\begin{tikzpicture}
			\begin{axis}[
			height = 4cm,
			width = 0.5\columnwidth,
			xlabel near ticks, 
			xlabel = {t [\si{\second}]},
			ylabel near ticks,  
			ylabel shift={-4pt},
			ylabel = {$\conn$ },
			ymin=0, ymax=3,
			ytick={0, 1, 2, 3},
			yticklabels = {$0$, $1$, $2$, $3$},
			xmin=0, xmax=60,
			legend image post style={only marks, mark=-},
			legend style={font=\footnotesize}
			]
			\addplot [line width=1pt, color=red] 
			table[x = x_pos, y = y_pos,col sep=space, row sep=\\]
			{
				x_pos	y_pos\\
				0  0.3\\
				60  0.3\\
			};
			\addplot [line width=1.4pt, color=color1] table [x index=1,y index=0,col sep=comma] {data/abs_vel_no_heu_real.csv};
			\legend{$\varepsilon = 0.3$}
			\end{axis}
		\end{tikzpicture}
	\label{fig:relative_real}}
	\end{subfloat}
	\caption{Algebraic connectivity in ideal and real conditions with disconnecting behavior. In the real conditions, the threshold $\varepsilon$ is crossed due to the presence of delay.}
	\label{fig:initial_guess}
\end{figure}

\subsection{Investigative simulations}
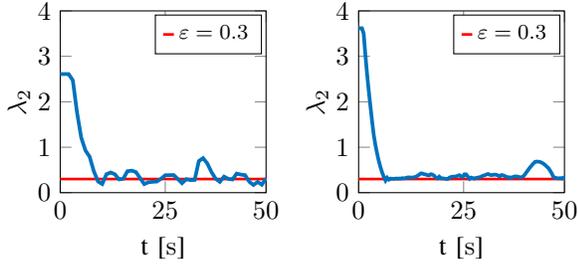
\begin{figure}[tb]
	\centering
	\subfloat[][\centering With delay addition \mbox{(delay = $0.1\si{\second}$,} variable delay = 0).]{
		\begin{tikzpicture}
			\begin{axis}[
				height = 4cm,
				width = 0.5\columnwidth,
				xlabel near ticks, 
				xlabel = {t [\si{\second}]},
				ylabel near ticks, 
				ylabel shift={-4pt},
				ylabel = {$\conn$ },
				ymin=0, ymax=4,
				ytick={0, 1, 2, 3, 4},
				yticklabels = {$0$, $1$, $2$, $3$, $4$},
				xmin=0, xmax=49,
				xtick={0,25, 49},
				xticklabels = {$0$, $25$, $50$},
				legend image post style={only marks, mark=-},
				legend style={font=\footnotesize}
				]
				\addplot [line width=1pt, color=red] table [x = x_pos, y = y_pos,col sep=space, row sep=\\]
				{
					x_pos y_pos\\
					0 0.3\\
					500 0.3\\
				};
				\addplot [line width=1.4pt, color=color1] table [x index=1,y index=0,col sep=comma] {data/abs_vel_no_heu_simulation.csv};	
				\legend{$\varepsilon = 0.3$}
			\end{axis}
		\end{tikzpicture}\label{fig:simulation_with_delay}} 
	\subfloat[][\centering Without delay addition \mbox{(delay = $0\si{\second}$,} variable delay = 0).]{
		\begin{tikzpicture}
			\begin{axis}[
				height = 4cm,
				width = 0.5\columnwidth,
				xlabel near ticks, 
				xlabel = {t [\si{\second}]},
				ylabel near ticks,  
				ylabel shift={-4pt},
				ylabel = {$\conn$ },
				ymin=0, ymax=4,
				ytick={0, 1, 2, 3, 4},
				yticklabels = {$0$, $1$, $2$, $3$, $4$},
				xmin=0, xmax=49,
				xtick={0,25, 49},
				xticklabels = {$0$, $25$, $50$},
				legend image post style={only marks, mark=-},
				legend style={font=\footnotesize}
				]
				\addplot [line width=1pt, color=red] table[x = x_pos, y = y_pos,col sep=space, row sep=\\]
				{
					x_pos y_pos \\
					0 0.3 \\
					500 0.3 \\
				};
				\addplot [line width=1.4pt, color=color1] table [x index=1,y index=0,col sep=comma] {data/simulation_without_delay.csv};
				\legend{$\varepsilon = 0.3$}
			\end{axis}
		\end{tikzpicture}\label{fig:simulation_without_delay}}
	\caption{Algebraic connectivity in the simulations for delay analysis with disconnecting behavior ($N=10$, $\varepsilon=0.3$).}
	\label{fig:delay_analysis}
\end{figure}

To understand which disturbing factor caused the decreased performance
of the control law, we reproduced the same experimental conditions
in simulation with ARGoS. 
In particular, we add a delay in the calculation of $\conn$ and of
$v_2$.
Fig.~\ref{fig:simulation_with_delay} reports the trend of $\conn$ in a
simulation with the disconnecting behavior as the desired behavior 
for the robots, and the similarity is clear with the trend in the physical robots
experiment (Fig,~\ref{fig:relative_real}). Instead,
Fig.~\ref{fig:simulation_without_delay} reports the trends of $\conn$ without
the addition of the delay, and the behavior is approaching the ideal case
reported in Fig.~\ref{fig:disconnecting_ideal_matlab}. The small differences
are probably caused by the different kinematics of the robots and the limited
input.

From these simulations we deduce that the main issue is the delay, which
corroborates the introduction of the heuristic proposed in
Section~\ref{sec:heuristic}.

\subsection{Experiments with the heuristic for delay} 
\label{sec:exp_delay}
In this section, we test the efficacy of the heuristic proposed
in~\eqref{eq:heuristic} for the disconnecting
behavior.

\subsubsection{Simulations}
\begin{figure}[tb]	
	\centering
	\begin{tikzpicture}
	\begin{axis}[
	height = 4cm,
	width = \columnwidth,
	xlabel near ticks, 
	xlabel = {t [\si{\second}]},
	ylabel near ticks, 
	ylabel shift={-4pt},
	ylabel = {$\conn$ },
	ymin=0, ymax=4,
	ytick={0, 1, 2, 3, 4},
	yticklabels = {$0$, $1$, $2$, $3$, $4$},
	xmin=0, xmax=49,
	xtick={0,25, 49},
	xticklabels = {$0$, $25$, $50$},
	cycle list name=MyCyclelist	,
	legend image post style={only marks, mark=-},
	legend style={font=\footnotesize}
	]
	\addplot [line width=1pt, color=red] 
	table[x = x_pos, y = y_pos]
	{
		x_pos	y_pos
		0  0.3
		500  0.3
	};
	\foreach \i in {0,1,...,19} {\addplot+  table [x index=20,y index=\i,col sep=comma] {data/simulation_heuristic_new.csv}; }	
	\legend{$\varepsilon = 0.3$}
	\end{axis}
	\end{tikzpicture}
	
	\caption{Algebraic connectivity in 20 simulations with the proposed heuristic and disconnecting behavior ($N=10$, \mbox{$\varepsilon = 0.3$}, delay = 0.05 \si{\second}, variable delay = 1).}
	\label{fig:simulation_heuristic_all}
\end{figure}
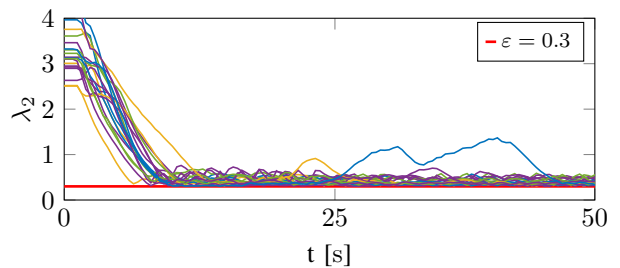

We investigated different combinations of variables: number of robots,
threshold, and value of delay (possibly different for each robot). The
diversification is aimed at verifying the effectiveness in a wide
range of cases. For each combination of the variables we performed 20 experiments with random
initial positions of the robots. Fig.~\ref{fig:simulation_heuristic_all}
reports the behavior of $\conn$ in every trial for one combination of the
variables.

Table~\ref{tab:simulations_heuristic} summarizes the results of the
simulations in an aggregate form. In particular, for each trial we recorded
the minimum value of $\conn$: the last three columns of
Table~\ref{tab:simulations_heuristic} report the mean value (\textit{Mean}),
the standard deviation (\textit{Std}), and the minimum value (\textit{Min}),
for each combination of the parameters, reported in the first four columns of
the table.
%
%
In particular, the third column (\textit{Max delay}) was the maximum delay
injected in the system. In addition, if the value in the fourth column
(\textit{Delay variable}) is equal to zero, then all the robots had the same
delay, if the value is one, the robots worked with random delay, uniformly chosen between the
maximum value and the null value (at least one robot had delay equal to the
maximum value).
\begin{table}[tb] 
	\centering
	\caption{Data of the simulations with the heuristic correction ($\kappa = 0.04$).}
	\label{tab:simulations_heuristic} 
	\begin{tabular}{ ccC{1 cm }C{1 cm}ccc } 
		\toprule
		$N$ & $\varepsilon$ &Delay max [s]&Delay variable& Mean & Std & Min\\
		\midrule
		\rowcolor[RGB]{204,255,255} 5 & 0.1 & 0 & 0 & 0.104 & 0.003 & 0.099 \\
		\midrule
		5 &	0.1 & 0.1 & 0	&	 	 	0.177 &	0.023 &	0.134 \\
		\midrule
	 	\rowcolor[RGB]{220,220,220} 5 &	0.1 & 0.1 & 1	&	 	 	0.126 &	0.027 &	0.081 \\
		\midrule
		5 &	0.1 & 0.05 & 0	&	 	 	0.126 &	0.013 & 0.109\\
		\midrule
		\rowcolor[RGB]{220,220,220} 5 &	0.1 & 0.05 & 1	&	 	 	0.123 & 0.079 & 0.082 \\
		\midrule
		\rowcolor[RGB]{204,255,255} 5 & 0.3 & 0 & 0 & 0.310 & 0.008 & 0.295 \\
		\midrule
		5 &	0.3 & 0.1 & 0	&	 	 	0.458 & 0.046 & 0.400\\
		\midrule
		\rowcolor[RGB]{220,220,220} 5 &	0.3 & 0.1 & 1	&	 	 	0.370 & 0.083 & 0.265 \\
		\midrule
		5 &	0.3 & 0.05	& 0	&	 	 	0.359 & 0.025 & 0.315 \\
		\midrule
		\rowcolor[RGB]{220,220,220} 5 & 0.3 & 0.05 & 1	&	 	 	0.321 & 0.026  & 0.277\\
		\midrule
		\rowcolor[RGB]{204,255,255} 10 & 0.1 & 0 & 0 & 0.103 & 0.005 & 0.088 \\
		\midrule
		10 & 0.1 & 	0.1 & 0 & 0.213 & 0.030 & 0.149 \\
		\midrule
		\rowcolor[RGB]{220,220,220} 10 & 0.1 & 	0.1 & 1 & 0.129 & 0.041 & 0.075 \\
		\midrule
		10 & 0.1 & 	0.05 & 0 & 0.136 & 0.015 & 0.114 \\
		\midrule
		\rowcolor[RGB]{220,220,220} 10 & 0.1 & 	0.05 & 1 & 0.105 & 0.016 & 0.078 \\
		\midrule
		\rowcolor[RGB]{204,255,255} 10 & 0.3 & 0 & 0 & 0.317 & 0.009 & 0.299 \\
		\midrule
		10 & 0.3 & 	0.1 & 0 & 0.603  & 0.059 & 0.492 \\
		\midrule
		\rowcolor[RGB]{220,220,220} 10 & 0.3 & 	0.1 & 1 & 0.433 & 0.100 & 0.272 \\
		\midrule
		10 & 0.3 & 	0.05 & 0 & 0.340 & 0.031 & 0.343 \\
		\midrule
		\rowcolor[RGB]{220,220,220} 10 & 0.3 & 	0.05 & 1 & 0.326 & 0.042 & 0.262 \\
		\bottomrule
	\end{tabular}
\end{table}

It is worth noting that in just few cases (highlighted in grey in
Table~\ref{tab:simulations_heuristic}) the minimum value of $\conn$ fell below
the given threshold, and this happened when using variable delays. However,
even in these particular cases, the proposed heuristic compensates the effect
introduced by the delay, and the minimum value of $\conn$ is comparable to the one obtained in absence
of delay (highlighted in light blue in Table~\ref{tab:simulations_heuristic}).

In the simulations without delay, the mean value is guaranteed to be above the threshold by the presence of the parameter $\kappa$, which allows to compensate the aforementioned other disturbing factors. During the simulations, the parameter $\kappa$ was empirically tuned for the
actual environment: we used the value \mbox{$\kappa=0.04$}, which provided good
results with all parameter combinations.

\subsubsection{Real experiments}
\begin{figure}[tb]	
	\centering
	\begin{tikzpicture}
		\begin{axis}[
			height = 4cm,
			width = \columnwidth,
			xlabel near ticks, 
			xlabel = {t [\si{\second}]},
			ylabel near ticks, 
			ylabel shift={-4pt},
			ylabel = {$\conn$ },
			ymin=0, ymax=3,
			ytick={0, 1, 2, 3},
			yticklabels = {$0$, $1$, $2$, $3$},
			xmin=0, xmax=70,
			xtick={0,35,70},
			xticklabels = {$0$, $35$, $70$},
			cycle list name=MyCyclelist	,
			legend image post style={only marks, mark=-},
			legend style={font=\footnotesize},
			legend columns = 2
			]
			\addplot [line width=1pt, color=red] 
			table[x = x_pos, y = y_pos]
			{
				x_pos	y_pos
				0  0.3
				500  0.3
			};
			\foreach \i in {0,1,...,3} {\addplot+  table [x index=4,y index=\i,col sep=comma] {data/heuristic_influence.csv}; }	
			\legend{$\varepsilon = 0.3$, Ground truth, Robot 1, Robot 2, Robot 3}
		\end{axis}
	\end{tikzpicture}
	
	\caption{Difference between the actual algebraic connectivity and the values used by the robots. ($N=3$, $\varepsilon = 0.3$).}
	\label{fig:heuristic_influence}
\end{figure}
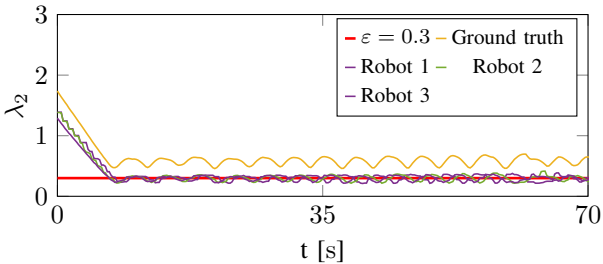

For testing the effectiveness of the proposed heuristic, we replicated
the experiments of the disconnecting behavior in the real setup
(described in Section~\ref{sec:hardware_software_implementation}). We
empirically tuned $\kappa$ at the value of 0.05, and we measured that
$\tau \approx 0.3$ \si{\second}. We performed several experiments with 3 robots starting in random initial
positions. Fig.~\ref{fig:real_heuristic} shows the value of $\conn$
for a representative trial, with starting positions of the robots
similar to the ones of the experiment reported in
Fig.~\ref{fig:relative_real}. With the introduction of the heuristic,
the threshold value is never exceeded, differently from what happens
without the correction for the presence of delay
(Fig.~\ref{fig:relative_real}). To further validate the efficacy of
the heuristic, we replicated the experiment with different starting
positions. Fig.~\ref{fig:real_heuristic_all} reports the behavior of
$\conn$ in the ten different trials.  The effect of the proposed
heuristic on the calculation of $\conn$ on the robots is highlighted
in Fig.~\ref{fig:heuristic_influence}. The actual value of $\conn$ is
calculated with the ground truth data, namely the data extrapolated
from the Optitrack system, which report the exact distance among the
robots. Instead, the value of $\conn$ calculated on the robots is
lower as a consequence of considering a worst-case distance.  Finally,
these experiments confirm the delay-compensation capacity of the
proposed heuristic.

\begin{figure}[tb]
	\centering
	\begin{subfloat}[][\centering Initial positions similar to experiment reported in Fig.~\ref{fig:relative_real}.]{
		\begin{tikzpicture}
		\begin{axis}[
		height = 4cm,
		width = 0.5\columnwidth,
		xlabel near ticks, 
		xlabel = {t [\si{\second}]},
		ylabel near ticks, 
		ylabel shift={-4pt},
		ylabel = {$\conn$ },
		ymin=0, ymax=3,
		ytick={0, 1, 2, 3, 4},
		yticklabels = {$0$, $1$, $2$, $3$, $4$},
		xmin=0, xmax=70,
		xtick={0,35, 70},
		xticklabels = {$0$, $35$, $70$},
		cycle list name=MyCyclelist	,
		legend image post style={only marks, mark=-},
		legend style={font=\footnotesize}
		]
		\addplot [line width=1pt, color=red] 
		table[x = x_pos, y = y_pos,col sep=space, row sep=\\]
		{
			x_pos	y_pos\\
			0  0.3\\
			500  0.3\\
		};
		\addplot [line width=1.4pt, color=color1] table [x index=1,y index=0,col sep=comma] {data/real_heuristic_single_new.csv};	
		\legend{$\varepsilon = 0.3$}
		\end{axis}
		\end{tikzpicture}
		\label{fig:real_heuristic}}
	\end{subfloat}
	\begin{subfloat}[][\centering Random initial positions (10 replications).]{
		\begin{tikzpicture}
		\begin{axis}[
		height = 4cm,
		width = 0.5\columnwidth,
		xlabel near ticks, 
		xlabel = {t [\si{\second}]},
		ylabel near ticks, 
		ylabel shift={-4pt},
		ylabel = {$\conn$ },
		ymin=0, ymax=3,
		ytick={0, 1, 2, 3},
		yticklabels = {$0$, $1$, $2$, $3$},
		xmin=0, xmax=70,
		xtick={0,35,70},
		xticklabels = {$0$, $35$, $70$},
		cycle list name=MyCyclelist	,
		legend image post style={only marks, mark=-},
		legend style={font=\footnotesize}
		]
		\addplot [line width=1pt, color=red] 
		table[x = x_pos, y = y_pos,col sep=space, row sep=\\]
		{
			x_pos	y_pos\\
			0  0.3\\
			500  0.3\\
		};
		\foreach \i in {0,1,...,9} {\addplot+  table [x index=10,y index=\i,col sep=comma] {data/real_heuristic_new.csv}; }	
		\legend{$\varepsilon = 0.3$}
		\end{axis}
		\end{tikzpicture}
		\label{fig:real_heuristic_all}}
	\end{subfloat}
	\caption{Algebraic connectivity in the experiments on real robots with the proposed heuristic and disconnecting behavior ($N=3$, $\varepsilon = 0.3$).}
	\label{fig:heuristic_experiments}
\end{figure}
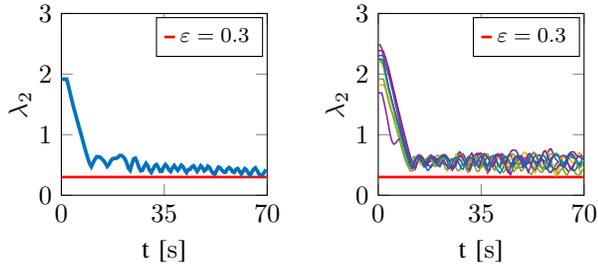

\subsection{Performance analysis}\label{sec:coveraga}
The second desired behavior is a coverage
task~\cite{cortes2004coverage}, a standard problem of multi-robot
systems, that we tested to investigate the effect on performance of
choosing the worst-case scenario in~\eqref{eq:heuristic}. We performed
a set of ten experiments, in which three Khepera IV robots used the
available neighbours' positions to calculate a Voronoi tessellation
using Fortune's algorithm~\cite{fortune1987sweepline}, and then each
robot carried out a Lloyd relaxation by chasing the center of its
cell, in a manner similar to~\cite{cortes2004coverage}.  The area to
be covered was $5.29\; \si{\m^2}$ and the maximum covered area
achievable was $3.68\; \si{\m^2}$, with the given threshold
$\varepsilon = 0.3$, and with a sensing range
$R_{sensing} = 0.75\; \si{\m}$. We need to refer to a particular
$\varepsilon$ because a different choice of $\varepsilon$ causes
different performance of the system. However, the analysis of the
influence of $\varepsilon$ on the performances of the system goes
beyond the scope of this paper.

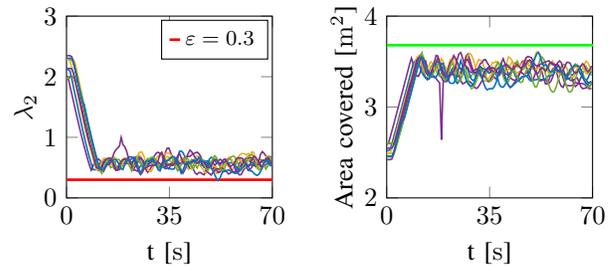
\begin{figure}[tb]
	\centering
	\begin{subfloat}[][\centering Algebraic connectivity during the experiments.]{
		\begin{tikzpicture}
			\begin{axis}[
			height = 4cm,
			width = 0.5\columnwidth,
			xlabel near ticks, 
			xlabel = {t [\si{\second}]},
			ylabel near ticks, 
			ylabel shift={-4pt},
			ylabel = {$\conn$ },
			ymin=0, ymax=3,
			ytick={0, 1, 2, 3},
			yticklabels = {$0$, $1$, $2$, $3$},
			xmin=0, xmax=70,
			xtick={0,35,70},
			xticklabels = {$0$, $35$, $70$},
			cycle list name=MyCyclelist	,
			legend image post style={only marks, mark=-},
			legend style={font=\footnotesize}
			]
			\addplot [line width=1pt, color=red] table[x = x_pos, y = y_pos,col sep=space, row sep=\\]
			{
				x_pos	y_pos\\
				0  0.3\\
				500  0.3\\
			};
			\foreach \i in {0,1,...,9} {\addplot+  table [x index=10,y index=\i,col sep=comma] {data/coverage_conn.csv}; }	
			\legend{$\varepsilon = 0.3$}
			\end{axis}
		\end{tikzpicture}
		\label{fig:coverage_conn}}
	\end{subfloat}
	\begin{subfloat}[][\centering Area covered with respect to the maximum area achievable with the chosen $\varepsilon$ (reported in \textit{green}).]{
		\begin{tikzpicture}
		\begin{axis}[
		height = 4cm,
		width = 0.5\columnwidth,
		xlabel near ticks, 
		xlabel = {t [\si{\second}]},
		ylabel near ticks, 
		ylabel shift={-4pt},
		ylabel = {Area covered [\si{\m^2}] },
		ymin=2, ymax=4,
		ytick={2, 3, 4},
		yticklabels = {$2$, $3$, $4$},
		xmin=0, xmax=70,
		xtick={0,35,70},
		xticklabels = {$0$, $35$, $70$},
		cycle list name=MyCyclelist	,
		]
		\addplot [line width=1pt, color=green] table[x = x_pos, y = y_pos,col sep=space, row sep=\\]
		{
			x_pos	y_pos\\
			0  3.68\\
			500  3.68\\
		};
		\foreach \i in {0,1,...,9} {\addplot+  table [x index=10,y index=\i,col sep=comma] {data/coverage_area.csv}; }	
		\end{axis}
		\end{tikzpicture}
		\label{fig:coverage_area}}
	\end{subfloat}
	\caption{Coverage experiments with $N=3$, $\varepsilon = 0.3$, $\kappa = 0.05$ (10 replications).}
	\label{fig:coverage}
\end{figure}

Hence, to analyse the impact of the proposed heuristic, we report, in
Fig.~\ref{fig:coverage_area}, the area covered during the experiments
with respect to the maximum achievable. It is clear how the maximum
value is never reached, but the performance is not drastically
decreased. Considering all the experiments, we obtain a mean deviation
of $12.71\%$ from the maximum area, with standard deviation equal to
$3.03\%$. In addition, it is important to remark that, without the
proposed heuristic, algebraic connectivity goes below the desired
value, and the robots usually disconnect.

\section{Conclusions}
\label{sec:conclusion}
In this paper we have presented a first implementation on real robots
of the proposed Control Barrier Function for connectivity maintenance,
first introduced
in~\cite{capelli2020cbf}. 
The experiments show that the CBF is effective in a more complex,
non-ideal, decentralized scenario, but with decreased performance. We found out that the main issue is the presence of delay in the system, in particular the fact that each robot works with a delayed
state of the system. To solve this problem we have proposed a
heuristic approach and we have demonstrated its effectiveness in
simulation and on real robots.

The proposed heuristic was tuned for the system and for the
experiments considered in Section~\ref{sec:exp_delay}, but it may give
an idea on how to deal with the presence of delay in a system when
using CBFs. As reported in~\cite{orosz2019safety}, computing
explicitly the safety domain in the presence of delays is
challenging. Hence, we propose a heuristic approach that maintains
comparable performance to the ideal case, as stated in
Section~\ref{sec:coveraga}. The proposed heuristic was developed
explicitly considering the system detailed in Section~\ref{sec:exp_delay}, but
the concept can be easily extended to more general scenarios, in which
the control law is computed based on relative positions among the
robots. In addition, the proposed heuristic does not increase the
computation time with respect to the nominal case, since the modified
inter-robot distance is trivially computed with constant factors.

It is worth noting that our implementation represents only an initial step
towards real-world implementation: the Optitrack system allows to obtain very
precise positions, and hence the relative measures among the robots are not
affected by noise, as they would be in a real application. Including noisy
measurements is part of our future work.

Moreover, besides the presence of delays, we considered an ideal communication
channel (no packet drops or link failures). As future work, we aim at
investigating the effect of these issues on the performance of the system.
Finally, to refine an algorithm for tuning the parameters of the proposed
heuristic, we want to investigate the effects of the specific communication
topologies on the information propagation in the presence of delays, and
possibly introduce  a machine learning based algorithm for tuning the
parameters.

\bibliographystyle{IEEEtran}
\bibliography{biblio}

\end{document}

%% file: images/heuristic.eps_tex
\begingroup%
  \makeatletter%
  \providecommand\color[2][]{%
    \errmessage{(Inkscape) Color is used for the text in Inkscape, but the package 'color.sty' is not loaded}%
    \renewcommand\color[2][]{}%
  }%
  \providecommand\transparent[1]{%
    \errmessage{(Inkscape) Transparency is used (non-zero) for the text in Inkscape, but the package 'transparent.sty' is not loaded}%
    \renewcommand\transparent[1]{}%
  }%
  \providecommand\rotatebox[2]{#2}%
  \ifx\svgwidth\undefined%
    \setlength{\unitlength}{1042.62289527bp}%
    \ifx\svgscale\undefined%
      \relax%
    \else%
      \setlength{\unitlength}{\unitlength * \real{\svgscale}}%
    \fi%
  \else%
    \setlength{\unitlength}{\svgwidth}%
  \fi%
  \global\let\svgwidth\undefined%
  \global\let\svgscale\undefined%
  \makeatother%
  \begin{picture}(1,0.75843168)%
    \put(0,0){\includegraphics[width=\unitlength]{heuristic.eps}}%
    \put(0.17640989,0.06125348){\color[rgb]{0,0,0}\makebox(0,0)[lb]{\smash{$v_{max}\left(\tau + \kappa\right) $}}}%
    \put(0.64618257,0.4154331){\color[rgb]{0,0,0}\makebox(0,0)[lb]{\smash{$d_{i,j}$}}}%
    \put(0.77885648,0.47919256){\color[rgb]{0,0,0}\makebox(0,0)[lb]{\smash{$\delta_{i,j}$}}}%
    \put(0.33815805,0.44009888){\color[rgb]{0,0,0}\makebox(0,0)[lb]{\smash{i}}}%
    \put(0.73568837,0.1343412){\color[rgb]{0,0,0}\makebox(0,0)[lb]{\smash{j}}}%
    \put(0.06412387,0.14523807){\color[rgb]{0,0,0}\makebox(0,0)[lb]{\smash{k}}}%
  \end{picture}%
\endgroup%